\pdfoutput=1
\documentclass[11pt]{article}
\usepackage[preprint]{acl}
\usepackage{times}
\usepackage{latexsym}
\usepackage[T1]{fontenc}
\usepackage[utf8]{inputenc}
\usepackage{microtype}
\usepackage{inconsolata}
\usepackage{graphicx}
\usepackage{multirow}
\usepackage{booktabs}
\usepackage{amsmath}
\usepackage{algorithm}
\usepackage{algorithmic}
\usepackage{amsfonts}
\usepackage{subcaption}

\usepackage{listings}
\usepackage{enumitem}
\usepackage{soul}
\definecolor{myyellow}{rgb}{1, 0.972549, 0.584313}
\definecolor{mygreen}{rgb}{0.772549, 0.945098, 0.756862}
\definecolor{myblue}{rgb}{0.811764, 0.866666, 0.996078}
\definecolor{myred}{rgb}{0.984313, 0.749019, 0.737254}
\definecolor{darkgreen}{rgb}{0, 0.5001960, 0}
\definecolor{darkred}{rgb}{0.8, 0, 0}

\title{PT-MoE: An Efficient Finetuning Framework for Integrating Mixture-of-Experts into Prompt Tuning}

\author{Zongqian Li, Yixuan Su, Nigel Collier \\
    University of Cambridge \\
    \texttt{\{zl510, ys484, nhc30\}@cam.ac.uk}}

\begin{document}
\maketitle
\begin{abstract}
Parameter-efficient fine-tuning (PEFT) methods have shown promise in adapting large language models, yet existing approaches exhibit counter-intuitive phenomena: integrating router into prompt tuning (PT) increases training efficiency yet does not improve performance universally; parameter reduction through matrix decomposition can improve performance in specific domains. Motivated by these observations and the modular nature of PT, we propose PT-MoE, a novel framework that integrates matrix decomposition with mixture-of-experts (MoE) routing for efficient PT. Results across 17 datasets demonstrate that PT-MoE achieves state-of-the-art performance in both question answering (QA) and mathematical problem solving tasks, improving F1 score by 1.49 points over PT and 2.13 points over LoRA in QA tasks, while enhancing mathematical accuracy by 10.75 points over PT and 0.44 points over LoRA, all while using 25\% fewer parameters than LoRA. Our analysis reveals that while PT methods generally excel in QA tasks and LoRA-based methods in math datasets, the integration of matrix decomposition and MoE in PT-MoE yields complementary benefits: decomposition enables efficient parameter sharing across experts while MoE provides dynamic adaptation, collectively enabling PT-MoE to demonstrate cross-task consistency and generalization abilities. These findings, along with ablation studies on routing mechanisms and architectural components, provide insights for future PEFT methods. \footnote{\url{https://github.com/ZongqianLi/PT-MoE}}
\end{abstract}

\section{Introduction}

Large language models (LLMs) have shown remarkable capabilities but require improvements in efficiency across data \citep{li2025auto}, training, and inference \citep{li2024500xcompressorgeneralizedpromptcompression, li-etal-2025-prompt, li2025reasongraphvisualisationreasoningpaths}. PEFT methods address training efficiency challenge by updating only a small subset of parameters \citep{han2024parameterefficient}. \textbf{Prompt tuning} (PT) stands out among PEFT approaches with its unique advantages: minimizing trainable parameters through soft prompt optimization, enabling modular utilization through task-specific prompts without model modifications, and supporting flexible knowledge composition \citep{lester-etal-2021-power}. These properties make it effective for low-resource and multi-task applications where efficient adaptation is essential.

\begin{figure}[t!]
    \centering
    \includegraphics[width=0.47\textwidth]{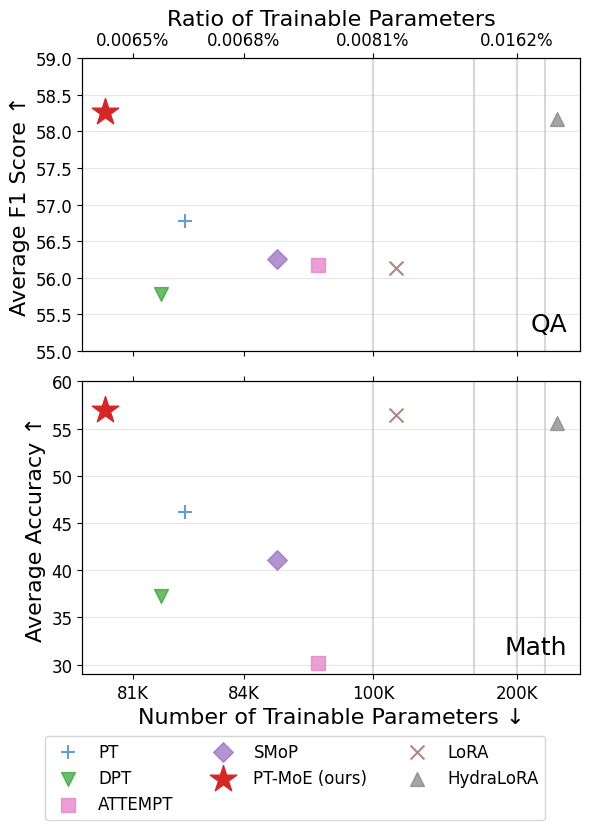}
    \caption{\textbf{Performance comparison} of PEFT methods on 12 QA datasets in the MRQA benchmark (upper) and 5 math datasets (lower). ↑ indicates higher is better; ↓ indicates lower is better.}
    \label{cover_figure}
\end{figure}

Despite these advantages, we observe two \textbf{counter-intuitive phenomena} in prompt tuning. First, integrating router into prompt tuning does not decrease training efficiency yet improves performance in specific domains rather than universally (SMoP vs PT in Table \ref{mrqa-f1}), suggesting domain-dependent optimization dynamics. \textbf{Second}, decomposing soft prompts into low-rank matrices, while reducing parameters, can surprisingly improve model performance in specific areas (DPT vs PT in Table \ref{math}). These phenomena indicate that the relationship between parameter efficiency and model effectiveness in prompt tuning is more nuanced than previously understood, motivating the need for a more sophisticated approach to prompt optimization.

Based on these observations, we propose a novel framework, \textbf{Prompt Tuning with Efficient Mixture-of-Experts (PT-MoE)}, that combines matrix decomposition with MoE routing. As shown in Figure \ref{cover_figure}, our approach not only achieves state-of-the-art performance, but also uses minimal trainable parameters and moderate training steps.

Our work makes three key \textbf{contributions}: 

\vspace{-7pt}

\begin{itemize}[left=0pt, itemsep=0pt, parsep=0pt]

\item \textbf{Novel finetuning framework:} We propose PT-MoE, integrating matrix decomposition with MoE for prompt tuning. Our framework achieves state-of-the-art performance with fewer parameters while outperforming either method alone, demonstrating their complementary benefits.

\item \textbf{Design dynamics:} We thoroughly analyze key variables influencing the performance of PT-MoE, including prompt length, expert count, trainable parameters, routing mechanisms, and model size. Findings provide design guidelines for future parameter-efficient tuning approaches.

\item \textbf{Comprehensive analysis:} We provide detailed empirical studies across diverse tasks, including QA and mathematical problem solving, establishing a basis for future work in efficient finetuning methods.

\end{itemize}

\vspace{-7pt}

The remainder of this paper is organized as follows: Section \ref{Related Work} reviews related work in prompt tuning, covering both direct tuning approaches and transfer learning methods. Section \ref{Methods} presents our PT-MoE framework, detailing the matrix decomposition strategy, dynamic router design, and training methodology. Section \ref{Experimental Design} describes our experimental design across QA and mathematical problem-solving tasks. Section \ref{Results} presents comprehensive results, including detailed ablation studies analyzing the influence of prompt length, parameter count, expert number, routing mechanisms, and model size, followed by efficiency analysis. Section \ref{Conclusions} concludes with key findings and future directions.

\section{Related Work}
\label{Related Work}

To contextualize our approach, we review existing prompt tuning methods, which fall into two categories: direct prompt tuning approaches focusing on architectural innovations, and transfer learning methods enabling cross-task knowledge sharing.

\textbf{Direct prompt tuning} methods have developed into four main branches: (1) General approaches that directly optimize prompt parameters, including Prompt Tuning that prepends trainable vectors to input while freezing the language model \citep{lester-etal-2021-power}, XPrompt that employs hierarchical structured pruning to identify and retain important prompt tokens \citep{ma-etal-2022-xprompt}, and P-Tuning v2 that introduces deep prompts across all transformer layers \citep{liu-etal-2022-p}; (2) Encoder-based methods that leverage additional modules, such as P-Tuning that incorporates an encoder to learn dependencies between continuous embeddings \citep{liu2023gptunderstands}, Residual Prompt Tuning (RPT) that employs a residual part with down/up-projection layers for stable optimization \citep{razdaibiedina-etal-2023-residual}, and Prefix Tuning that prepends trainable key-value pairs at each layer through a reparameterization section \citep{li-liang-2021-prefix}; (3) Decomposition methods that decompose prompt embeddings, including Decomposed Prompt Tuning (DPT) that applies low-rank matrix decomposition to reduce parameter count \citep{xiao-etal-2023-decomposed}, and DePT that combines shorter soft prompts with low-rank updates to word embeddings \citep{shi2024dept}; and (4) MoE approaches such as Sparse Mixture-of-Prompts (SMoP) that employs multiple shorter prompts with a dynamic router to route inputs to the most suitable soft prompt \citep{choi2023smop}.

\textbf{Transfer learning} approaches in prompt tuning have developed into three categories: (1) General approaches that directly transfer prompt knowledge, including SPoT that introduces both generic transfer through multi-task pre-training and targeted transfer via task similarity matching \citep{vu-etal-2022-spot}, and ATTEMPT that dynamically combines multiple source prompts through an attention-based mixing mechanism with instance-level adaptation \citep{asai-etal-2022-attempt}; (2) Encoder-based methods that facilitate knowledge transfer through additional modules, such as TransPrompt that employs parallel task-specific and universal encoders with balancing mechanisms for obtaining both task-dependent and task-agnostic knowledge \citep{wang-etal-2021-transprompt}, and Cross-Task Prompt Tuning (CTPT) that leverages multi-head attention for cross-task knowledge transfer with dimension reduction and derivative-free optimization \citep{xu-etal-2023-efficient}; and (3) Decomposition methods exemplified by Multitask Prompt Tuning (MPT) that decomposes prompts into shared and task-specific components through knowledge distillation, enabling efficient transfer while preserving task-specific adaptability through a rank-one decomposition strategy \citep{wang2023multitask}.

\section{Methods}
\label{Methods}

\begin{figure}[t!]
    \centering
    \includegraphics[width=0.4\textwidth]{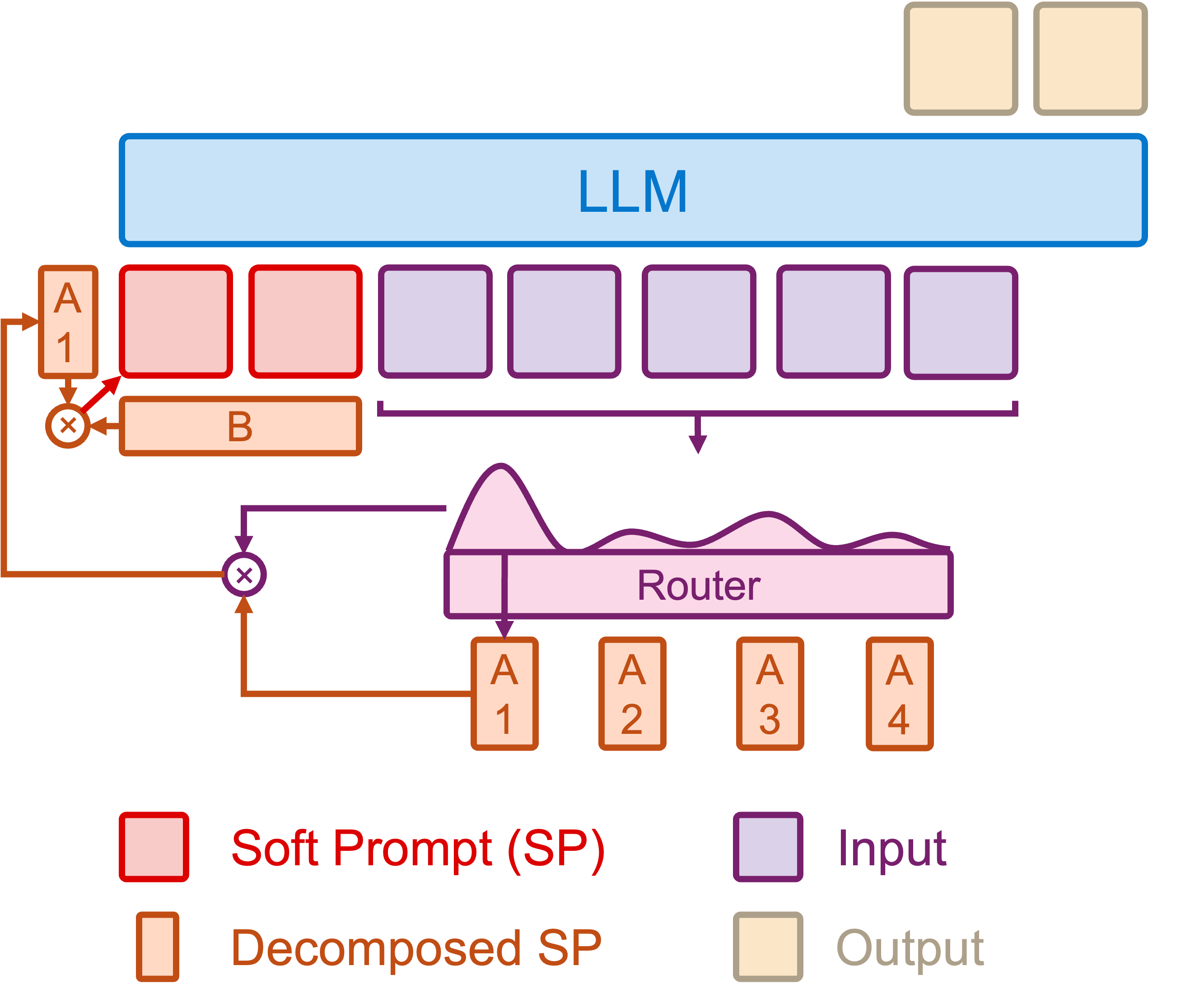}
    \caption{\textbf{Framework} of PT-MoE. Each soft prompt is decomposed into an input-specific matrix $A_i$ and a shared matrix $B$, with a router adaptively selecting and combining prompt components based on input. The resulting soft prompt is prepended to the input for the frozen LLM.}
    \label{method}
\end{figure}

Building upon the insights from prior work, we propose a new parameter-efficient prompt tuning framework, PT-MoE, shown in Figure \ref{method} and Algorithm \ref{pseudocode}.

\textbf{Framework Overview.} PT-MoE integrates matrix decomposition and dynamic routing. Given an input sequence $\mathbf{x}$, our framework first generates routing weights $\mathbf{w}$ through a router $R$: $\mathbf{w} = R(\mathbf{x})$. These weights determine the selection among $N$ decomposed prompts, where each prompt $\mathbf{P}_i$ is decomposed as $\mathbf{P}_i = \mathbf{A}_i\mathbf{B}$, with $\mathbf{A}_i$ being prompt-specific and $\mathbf{B}$ being shared across all prompts. The final soft prompt $\mathbf{P}$ is computed as $\mathbf{P} = \sum_{i=1}^{N} w_i\mathbf{A}_i\mathbf{B}$, which is then prepended to the input sequence for the frozen language model.

\begin{algorithm}[t!]
\small
\caption{Pseudo code of PT-MoE}
\begin{algorithmic}[1]
\REQUIRE Base model $\mathcal{M}$; input batch $X = {x_1,...,x_b}$; parameters $\theta$ \\
\hspace{-0.51cm}\textbf{Notation:} $b$ - batch size; $s$ - sequence length; $n$ - number of prompts; $k$ - tokens per prompt; $d$ - low-rank dimension; $h$ - hidden dimension \\
\FOR{batch $x \in X$}
\STATE Get input embeddings $E = \mathcal{M}_{\text{embed}}(x)$ \\
\textit{where} $E \in \mathbb{R}^{b \times s \times h}$ \\
\STATE Calculate mean embeddings \\
$\mu = \text{mean}(E, \text{dim}=1)$ \textit{where} $\mu \in \mathbb{R}^{b \times h}$ \\
\STATE Compute router logits $l = W\mu + b$ \\
\textit{where} $W \in \mathbb{R}^{n \times h}, b \in \mathbb{R}^{n}, l \in \mathbb{R}^{b \times n}$ \\
\STATE Get router weights \\
$w = \text{softmax}(l)$ \textit{where} $w \in \mathbb{R}^{b \times n}$
\FOR{each sample $j$ in batch}
\STATE Find indices of top-k weights: \\ $i_{topk} = \text{argsort}(w_j)[-k:]$ \\
\STATE Zero all weights except top-k: \\
$w_j[i] = 0$ for all $i \notin i_{topk}$ \\
\ENDFOR
\STATE Initialize prompt embeddings \\
$P = 0, P \in \mathbb{R}^{b \times k \times d}$ \\
\FOR{each weight $w_i$ in $w$}
\STATE Compute weighted prompts \\
$P = P + w_i A_i$ \textit{where} $A_i \in \mathbb{R}^{k \times d}$ \\
\ENDFOR
\STATE Project to model dimension \\
$P = P \times B$ \textit{where} $B \in \mathbb{R}^{d \times h}$ \\
\STATE Combine with input: $C = \text{concat}(P, E)$ \\
\textit{where} $C \in \mathbb{R}^{b \times (k+s) \times h}$ \\
\STATE Generate through base model: $y = \mathcal{M}(C)$ \\
\ENDFOR
\ENSURE Model predictions $y$ 
\end{algorithmic}
\label{pseudocode}
\end{algorithm}

\begin{table*}[t!]
\centering
\small
\begin{tabular}{p{0.1\textwidth}p{0.80\textwidth}}
\hline
\multicolumn{2}{l}{\textbf{MRQA (Extractive QA)}} \\
In-domain & SQuAD \citep{rajpurkar-etal-2016-squad}, TriviaQA \citep{joshi-etal-2017-triviaqa}, SearchQA \citep{dunn2017searchqanewqadataset}, HotpotQA \citep{yang-etal-2018-hotpotqa}, NaturalQuestions \citep{kwiatkowski-etal-2019-natural} \\
Out-of-domain & BioASQ \citep{partalas2013results}, DROP \citep{dua-etal-2019-drop}, DuoRC \citep{saha-etal-2018-duorc}, RACE \citep{lai-etal-2017-race}, RelationExtraction \citep{levy2017zero}, TextbookQA \citep{8100054} \vspace{0.5em} \\
\hline
\multicolumn{2}{l}{\textbf{Mathematics (Problem Solving)}} \\
In-domain & GSM8K \citep{cobbe2021trainingverifierssolvemath} \\
Out-of-domain & SVAMP: Subtraction, Addition, Common-Division, Multiplication \citep{patel-etal-2021-nlp}; ASDIV \citep{miao-etal-2020-diverse}; MAWPS \citep{koncel-kedziorski-etal-2016-mawps}; MATH\_PROBLEMS \citep{math-problems} \\
\hline
\end{tabular}
\caption{Overview of training and evaluation \textbf{datasets} that span two task categories: extractive QA (MRQA benchmark with 12 QA datasets) and mathematical problem solving (GSM8K and specific mathematical datasets). For each category, datasets are divided into in-domain sets used for training, validation, and evaluation, and out-of-domain sets used exclusively for testing generalization ability.}
\label{tab:datasets}
\end{table*}

\textbf{Matrix Decomposition.} To achieve parameter efficiency, we decompose each prompt matrix $\mathbf{P}_i \in \mathbb{R}^{T \times H}$ into a prompt-specific matrix $\mathbf{A}_i \in \mathbb{R}^{T \times R}$ and a shared matrix $\mathbf{B} \in \mathbb{R}^{R \times H}$, where $T$, $H$, and $R$ denote the prompt length, hidden dimension, and low-rank dimension respectively. This reduces parameters from $O(NTH)$ to $O(NTR + RH)$ for $N$ prompts. The low-rank dimension $R$ is either manually determined or computed to maintain parameter efficiency. For initialization, we first transform task-relevant text into word embeddings $\mathbf{E} \in \mathbb{R}^{T \times H}$, then perform SVD: $\mathbf{E} = \mathbf{U}\mathbf{\Sigma}\mathbf{V}^{\top}$. Each $\mathbf{A}_i$ is initialized as $\mathbf{U}{:R}\mathbf{\Sigma}{R}^{1/2}$ and the shared $\mathbf{B}$ as $\mathbf{\Sigma}{R}^{1/2}\mathbf{V}_{R:}^{\top}$, where subscript $R$ indicates truncation to the first $R$ components. This approach ensures the initial prompts have task-relevant information while maintaining the parameter efficiency of decomposition.

\textbf{Dynamic Router.} The router adaptively selects prompts based on input context. Given an input sequence embedding $\mathbf{x} \in \mathbb{R}^{H}$ (obtained by averaging token embeddings), the router computes logits through a linear projection: $\mathbf{l} = \mathbf{W}\mathbf{x} + \mathbf{b}$, where $\mathbf{W} \in \mathbb{R}^{N \times H}$ and $\mathbf{b} \in \mathbb{R}^{N}$. During training, we apply multiplicative Gaussian noise to encourage exploration: $\mathbf{l}' = \mathbf{l} \odot (1 + \epsilon)$, where $\epsilon \sim \mathcal{N}(0, \sigma^2)$. The routing weights are computed as $\mathbf{w} = \text{softmax}(\mathbf{l}') \odot \mathbf{1}_{\operatorname{argmax}}$, where $\mathbf{1}_{\operatorname{argmax}}$ is a one-hot vector with 1 at the position of the maximum value. This hard selection strategy reduces overlap between prompts while maintaining end-to-end differentiability through straight-through estimation.

\textbf{Training and Prediction.} During training, we optimize both the router parameters and decomposed prompt matrices while keeping the base model frozen. For language model training, we use negative log-likelihood loss computed only on non-prompt positions using a binary mask: $\mathcal{L} = -\sum_{t \in \mathcal{M}} \log p(y_t|x_{<t})$, where $\mathcal{M}$ denotes non-prompt positions. We employ AdamW optimizer with warmup followed by a constant learning rate schedule, and gradient accumulation for stable optimization. At inference, noise is not added in the router, ensuring deterministic prompt selection.

\section{Experimental Design}
\label{Experimental Design}

\subsection{Datasets}

We complete evaluations across \textbf{17} diverse datasets, as shown in Table \ref{tab:datasets}, where in-domain datasets are split into training, validation, and test sets, while out-of-domain datasets are used exclusively for testing. For \textbf{QA}, we utilize 12 MRQA datasets \citep{fisch2019mrqa}, with in-domain sets like SQuAD \citep{rajpurkar-etal-2016-squad} testing information extraction abilities and out-of-domain sets like DROP \citep{dua-etal-2019-drop} evaluating domain adaptation. For mathematical \textbf{problem solving}, we use GSM8K \citep{cobbe2021trainingverifierssolvemath} from MetaMath \citep{yu2024metamath} as our in-domain dataset, complemented by specific out-of-domain datasets including all the subsets of SVAMP \citep{patel-etal-2021-nlp}, ASDIV \citep{miao-etal-2020-diverse}, MAWPS \citep{koncel-kedziorski-etal-2016-mawps}, and MATHPROBLEMS \citep{math-problems}.

\subsection{Gold Standard and Baselines}

We employ full model fine-tuning as our \textbf{gold standard}, which updates all parameters but requires substantial computational resources. Our \textbf{baselines}\footnote{All methods are controlled to have similar parameter budgets, with detailed configurations shown in Table \ref{method-parameters} of the Appendix.} include typical methods from \textbf{prompt tuning categories}: For \textbf{direct prompt tuning}, we select (1) PT from general approaches, (2) DPT from decomposition methods, and (3) SMoP from MoE approaches. While \textbf{transfer learning} methods like (4) ATTEMPT typically involve multi-turn training, we also evaluate its architecture under similar training for comprehensive comparison. We additionally compare \textbf{other PEFT methods} including (5) LoRA and (6) HydraLoRA, with HydraLoRA adopting a MoE-like architecture that uses a shared down-projection matrix and multiple routed up-projection matrices. \textbf{These two LoRA-based methods require model architecture modifications unlike the modular nature of prompt tuning methods.}

\subsection{Evaluation Metrices}

We employ task-specific evaluation metrics. For extractive QA tasks from MRQA, we adopt two metrics: \textbf{F1} score, which evaluates the token-level overlap between predicted and ground truth answer spans, balancing precision and recall; and \textbf{Exact Match (EM)}, which measures the percentage of predictions that exactly match the ground truth. For mathematical problem solving tasks, we use \textbf{accuracy}, defined as the percentage of correctly solved problems with exact answer matches.

\begin{table*}[t!]
\centering
\scriptsize
\begin{tabular}{ccccccccccccccc}
\hline
\multirow{2}{*}{\textbf{Method}} & \textbf{\#} & \multicolumn{6}{c}{\textbf{In-domain}} & \multicolumn{6}{c}{\textbf{Out-of-domain}} & \multirow{2}{*}{\textbf{Avg.}} \\
& \textbf{para.} & SQ & News & Tri & Srch & HP & NQ & \multicolumn{1}{|c}{BSQ} & DP & DRC & RC & RE & TB & \\
\hline
FT & 1.2B & 78.76 & 48.69 & 71.04 & 71.35 & 72.96 & 67.56 & 70.19 & 43.87 & 48.11 & 43.44 & 81.60 & 52.71 & 62.52 \\
\hline
LoRA & 106k & 69.82 & 39.91 & 70.61 & 55.56 & 63.29 & 65.92 & 65.38 & 35.25 & 43.69 & 38.04 & 74.09 & 52.00 & 56.13 \\
HydraLoRA & 278k & 74.24 & 44.05 & 71.38 & 60.13 & 64.02 & 66.31 & 68.76 & 34.38 & 44.36 & 40.00 & 77.97 & 52.44 & 58.17 \\
\hline
PT & 81k & 72.31 & 48.18 & 65.93 & 49.74 & 58.69 & 62.18 & 68.59 & 40.39 & 43.30 & 42.10 & 82.43 & 47.34 & 56.77 \\
DPT & 81k & 70.99 & 48.42 & 65.41 & 46.94 & 58.49 & 61.65 & 65.56 & 38.80 & 43.64 & 41.89 & 80.85 & 46.62 & 55.77 \\
SMoP & 86k & 74.15 & \textbf{48.96} & 66.13 & 41.08 & 58.96 & 61.17 & 68.59 & 39.92 & 42.07 & 42.34 & 83.73 & \textbf{47.85} & 56.25 \\
ATTEMPT & 90k & \textbf{74.22} & 48.18 & 65.31 & 37.64 & 60.18 & 59.59 & 66.69 & 45.32 & 42.86 & \textbf{43.01} & \textbf{84.11} & 46.91 & 56.17 \\
\textbf{PT-MoE} & \textbf{80k} & 73.85 & 48.24 & \textbf{67.34} & \textbf{51.33} & \textbf{62.16} & \textbf{62.95} & \textbf{69.33} & \textbf{48.02} & \textbf{43.96} & 42.51 & 83.70 & 45.71 & \textbf{58.26} \\
\hline
\end{tabular}
\caption{Evaluation results (F1 scores) for various PEFT methods on MRQA datasets. SQ: SQuAD; News: NewsQA; Tri: TriviaQA; Srch: SearchQA; HP: HotpotQA; NQ: NaturalQuestions; BSQ: BioASQ; DP: DROP; DRC: DuoRC; RC: RACE; RE: RelationExtraction; TB: TextbookQA. The bold values indicate the best performance among prompt tuning-based methods.}
\label{mrqa-f1}
\end{table*}

\begin{table*}[t!]
\centering
\scriptsize
\begin{tabular}{ccccccccccccccc}
\hline
\multirow{2}{*}{\textbf{Method}} & \textbf{\#} & \multicolumn{6}{c}{\textbf{In-domain}} & \multicolumn{6}{c}{\textbf{Out-of-domain}} & \multirow{2}{*}{\textbf{Avg.}} \\
& \textbf{para.} & SQ & News & Tri & Srch & HP & NQ & \multicolumn{1}{|c}{BSQ} & DP & DRC & RC & RE & TB & \\
\hline
FT & 1.2B & 65.28 & 32.76 & 62.29 & 61.50 & 56.19 & 50.45 & 49.06 & 32.26 & 38.84 & 29.52 & 66.99 & 43.71 & 49.07 \\
\hline
LoRA & 106k & 56.26 & 25.26 & 64.11 & 46.10 & 47.48 & 49.54 & 42.02 & 25.48 & 33.24 & 24.92 & 58.58 & 44.17 & 43.09 \\
HydraLoRA & 278k & 61.63 & 27.80 & 64.32 & 50.06 & 47.73 & 49.59 & 44.01 & 24.75 & 33.57 & 26.11 & 62.68 & 43.97 & 44.69 \\
\hline
PT & 81k & 61.25 & 32.62 & 59.49 & 42.40 & 44.45 & 47.28 & 51.79 & 30.60 & 34.64 & 29.82 & 72.45 & 39.52 & 45.52 \\
DPT & 81k & 58.49 & \textbf{32.88} & 58.56 & 39.65 & 44.33 & 46.54 & 49.46 & 28.74 & 35.64 & 30.26 & 70.48 & 38.72 & 44.48 \\
SMoP & 86k & 63.15 & 32.81 & 59.48 & 34.51 & 43.80 & 46.39 & 50.06 & 29.94 & 34.11 & 30.56 & 74.59 & \textbf{40.25} & 44.97 \\
ATTEMPT & 90k & \textbf{63.71} & 32.50 & 58.71 & 31.24 & 45.77 & 45.66 & 49.26 & 36.06 & 34.84 & 30.41 & \textbf{75.13} & 39.52 & 45.23 \\
\textbf{PT-MoE} & \textbf{80k} & 63.34 & 32.85 & \textbf{60.87} & \textbf{43.98} & \textbf{47.29} & \textbf{48.18} & \textbf{52.06} & \textbf{37.12} & \textbf{35.64} & \textbf{31.75} & 74.18 & 38.25 & \textbf{47.13} \\
\hline
\end{tabular}
\caption{Evaluation results (Exact Match) for MRQA datasets.}
\label{mrqa-em}
\end{table*}

\subsection{Models}

We get our main results using \textbf{LLaMA-3.2-1B-Instruct} as the base model for fine-tuning methods \citep{grattafiori2024llama3herdmodels}. For ablation studies on model size, we additionally employ \textbf{LLaMA-3.2-3B-Instruct}.  

\section{Results}
\label{Results}

\subsection{Question Answering}

The results on MRQA datasets shown in Table \ref{mrqa-f1} and \ref{mrqa-em} demonstrate the effectiveness of PT-MoE across various QA tasks. We highlight seven key findings: 
(1) PT-MoE achieves superior overall performance with an average F1 score of 58.26\%, outperforming SMoP (56.25\%) by 2.01 points and the standard PT (56.77\%) by 1.49 points, establishing a new state-of-the-art on the MRQA benchmark. 
(2) This improvement is further validated by Exact Match metrics, where PT-MoE demonstrates even more gains (47.13\% for average, outperforming SMoP and PT by 2.16 and 1.61 points respectively).
(3) PT-MoE exhibits strong generalization abilities across both in-domain and out-of-domain scenarios. It achieves the highest performance on four out of six in-domain datasets and three out of six out-of-domain datasets. 
(4) The stability of PT-MoE is evidenced by consistent improvements over PT across 11 out of 12 datasets, with only marginal decreases in the RACE dataset. In contrast, SMoP shows performance decrease on 5 datasets compared to PT. 
(5) Individual architectural components show limited gains: both matrix decomposition (DPT, 55.77\% F1) and MoE (SMoP, 56.25\% F1) underperform standard prompt tuning (PT, 56.77\% F1).
(6) PT-MoE's integration of matrix decomposition and MoE yields complementary benefits, outperforming both DPT and SMoP by 2.49 and 2.01 points for F1 respectively. This improvement over individual approaches proves the mutually beneficial nature of these methods.
(7) Notably, while PT-MoE achieves lower overall performance than FT, it reaches comparable or even higher scores than FT on specific datasets such as DROP (48.02\% vs 43.87\% F1) while using only 80K parameters compared to FT's 1.2B.
These results collectively validate the effectiveness of the architectural design of PT-MoE and demonstrate its superior performance in accuracy and generalization across diverse QA scenarios.

\begin{table*}[t!]
\centering
\scriptsize
\begin{tabular}{cccccccccccc}
\hline
\multirow{2}{*}{\textbf{Method}} & \textbf{\#} & \multicolumn{1}{c}{\textbf{In-domain}} & \multicolumn{8}{c}{\textbf{Out-of-domain}} & \multirow{2}{*}{\textbf{Average}} \\
 & \textbf{para.} & GSM8K & \multicolumn{1}{|c}{Subtraction} & Addition & Division & Multiplication & SVAMP & ASDIV & MAWPS & MP500 & \\
\hline
FT & 1.2B & 58.15 & 68.75 & 64.40 & 62.50 & 48.48 & 61.03 & 86.04 & 82.53 & 30.60 & 63.67 \\
\hline
LoRA & 106k & 41.77 & 67.50 & 61.01 & 52.08 & 33.33 & 53.48 & 73.42 & 70.70 & 43.00 & 56.47 \\
HydraLoRA & 278k & 41.31 & 57.50 & 62.71 & 52.08 & 39.39 & 52.92 & 74.08 & 76.05 & 33.40 & 55.55 \\
\hline
PT & 81k & 34.11 & 41.87 & 50.84 & 66.66 & 33.33 & 48.18 & 60.13 & 57.18 & 31.20 & 46.16 \\
Decomp. PT & 81k & 26.08 & 43.12 & 35.59 & 64.58 & 27.27 & 42.64 & 56.14 & 43.09 & 18.20 & 37.23 \\
SMoP & 86k & 27.97 & 38.12 & 35.59 & 33.33 & 33.33 & 35.09 & 49.50 & 65.91 & 26.80 & 41.05 \\
ATTEMPT & 90k & 27.36 & 40.00 & 35.59 & 37.50 & 27.27 & 35.09 & 24.91 & 49.01 & 14.60 & 30.19 \\
\textbf{PT-MoE} & \textbf{80k} & \textbf{35.63} & \textbf{55.62} & \textbf{55.93} & \textbf{79.16} & \textbf{36.36} & \textbf{56.77} & \textbf{77.74} & \textbf{71.83} & \textbf{42.60} & \textbf{56.91} \\
\hline
\end{tabular}
\caption{Accuracy (\%) on \textbf{mathematical problem-solving} tasks with the number of trainable parameters shown in the second column. The first four out-of-domain datasets are from the SVAMP dataset. MP500 denotes the first 500 questions from MATH\_PROBLEMS.}
\label{math}
\end{table*}

\subsection{Mathematical Problem Solving}
The results on mathematical tasks reveal several characteristics compared to QA tasks. We highlight six key findings:
(1) PT-MoE achieves state-of-the-art performance with an average accuracy of 56.91\%, improving upon PT (46.16\%) by 10.75 points, demonstrating its effectiveness in mathematical reasoning.
(2) The benefits of MoE integration shows method-dependent characteristics: in prompt tuning approaches, PT-MoE and SMoP show different changes over PT (by +10.75 and -5.11 points respectively); when applied to LoRA methods, HydraLoRA shows slightly performance decrease compared to LoRA.
(3) LoRA-based methods demonstrate advantages in mathematical tasks compared to their performance in QA. While LoRA underperformed PT by 5.36 points in MRQA, it outperforms PT by 10.31 points in mathematical tasks, indicating task-specific strengths of different PEFT approaches.
(4) PT-MoE demonstrates unique cross-task consistency: while prompt tuning methods excel in QA tasks and LoRA-based methods in mathematical tasks, PT-MoE achieves the highest average performance in both domains, indicating robust adaptability across different problem types.
(5) While PEFT methods consistently underperform full fine-tuning, the performance gap is larger in mathematical tasks compared to QA tasks, with a wider performance range among different methods. Notably, PT-MoE achieves comparable or higher performance to full fine-tuning on specific datasets such as Division and MP500.
(6) PT-MoE demonstrates superior parameter efficiency, achieving higher performance than LoRA while using only 75\% of its parameters (80k vs 106k), and outperforming HydraLoRA which uses 3.5 times more parameters.
These findings highlight both the unique challenges of mathematical tasks and the robust adaptability of PT-MoE across different problem domains.

\begin{table*}[t!]
\centering
\tiny
\begin{tabular}{p{0.3\textwidth} p{0.3\textwidth} p{0.3\textwidth}}
\hline
\multicolumn{3}{p{0.9\textwidth}}{\textbf{Question:} Let $f(y) = y^4 -3y^3 +y - 3$ and $g(y) = y^3 + 7y^2 -2$. Find $f(y) + g(y)$.  Write your answer as a polynomial with terms of decreasing degree. \textbf{Standard Answer:} $y^4-2y^3+7y^2+y-5$} \\
\hline
\multicolumn{1}{p{0.3\textwidth}}{\textbf{Original:}

\#\# Step 1: Identify the polynomials f(y) and g(y)

The polynomial f(y) is given as $f(y) = y^4 -3y^3 +y - 3$ and the polynomial g(y) is given as $g(y) = y^3 + 7y^2 -2$.

\#\# Step 2: Add the polynomials f(y) and g(y)

To find the sum of f(y) and g(y), we need to add the corresponding terms of each polynomial.

\#\# Step 3: Perform the addition

$f(y) + g(y) = (y^4 -3y^3 +y - 3) + (y^3 + 7y^2 -2)$

\#\# Step 4: Combine like terms

Combine the terms with the same degree to simplify the expression.

\#\# Step 5: Simplify the expression

$f(y) + g(y) = y^4 -3y^3 + y^3 + 7y^2 - 2 \textcolor{orange}{- 3}$

\#\# Step 6: Combine like terms

Combine the terms with the same degree.

\#\# Step 7: Simplify the expression

$f(y) + g(y) = y^4 -2y^3 + 7y^2 \textcolor{red}{-2}$

\#\# Step 8: Write the final answer

The final answer is: $y^4 -2y^3 + 7y^2 \textcolor{red}{-2}$
} & 
\multicolumn{1}{p{0.3\textwidth}}{\textbf{Prompt Tuning:}

\#\# Step 1:  To find the sum of the two polynomials $f(y)$ and $g(y)$, we need to add their corresponding terms.

\#\# Step 2:  The polynomial $f(y) = y^4 - 3y^3 + y - 3$ has terms of degree 4, 3, 2, and 1, while the polynomial $g(y) = y^3 + 7y^2 - 2$ has terms of degree 3, 2, and 1.

\#\# Step 3:  When we add these two polynomials, we need to combine like terms, which means adding the coefficients of terms with the same degree.

\#\# Step 4:  The terms with degree 4 are $y^4$ and $\textcolor{blue}{y^4}$, so their sum is $\textcolor{red}{2y^4}$.

\#\# Step 5:  The terms with degree 3 are $-3y^3$ and $\textcolor{blue}{-3y^3}$, so their sum is $\textcolor{red}{-6y^3}$.

\#\# Step 6:  The terms with degree 2 are $\textcolor{red}{y}$ and $7y^2$, so their sum is $7y^2$.

\#\# Step 7:  The terms with degree \textcolor{red}{1} are $\textcolor{blue}{1}$ and $\textcolor{blue}{-2}$, so their sum is $\textcolor{red}{-1}$.

\#\# Step 8:  Therefore, the sum of the two polynomials $f(y)$ and $g(y)$ is $\textcolor{red}{2y^4 - 6y^3} + 7y^2 \textcolor{red}{- 1}$.

\#\# Step 9:  The answer is $\textcolor{red}{2y^4 - 6y^3} + 7y^2 \textcolor{red}{- 1}$.

} & 
\multicolumn{1}{p{0.3\textwidth}}{\textbf{PT-MoE:} 

\#\# Step 1: Identify the polynomials f(y) and g(y)

The given polynomials are $f(y) = y^4 -3y^3 +y - 3$ and $g(y) = y^3 + 7y^2 -2$.

\#\# Step 2: Add the polynomials f(y) and g(y)

To find $f(y) + g(y)$, we need to add the corresponding terms of the two polynomials.

\#\# Step 3: Perform the addition

$f(y) + g(y) = (y^4 -3y^3 +y - 3) + (y^3 + 7y^2 -2)$

\#\# Step 4: Combine like terms

Combine the terms with the same degree of y: $y^4 + (y^3 -3y^3) + (y^3 + 7y^2) + (y - 3) -2$

\#\# Step 5: Simplify the expression

Simplify the expression by combining the coefficients of like terms: $y^4 -2y^3 + 7y^2 + y - 5$

\#\# Step 6: Write the final expression

The expression $f(y) + g(y)$ is $y^4 -2y^3 + 7y^2 + y - 5$.

\#\# Note: $y$ is the variable, and the answer is $y^4 -2y^3 + 7y^2 + y - 5$.} \\
\hline
\end{tabular}
\caption{A \textbf{case study} comparing three solution approaches (Original, Prompt Tuning, and PT-MoE) for a polynomial addition problem. Errors in the outputs are highlighted in \textcolor{red}{red} (incorrect terms), \textcolor{orange}{orange} (missing terms), and \textcolor{blue}{blue} (hallucinated terms).}
\label{example-table}
\end{table*}

\subsection{Case Study}

To better understand the performance characteristics of PT-MoE, we present a detailed case study of polynomial addition in Table \ref{example-table}. In this example, the response of the base model exhibits information loss, specifically omitting the linear term during simplification steps, leading to an incorrect final result. The conventional prompt tuning approach exhibits hallucinations and conceptual errors, particularly in degree identification and term combination, resulting in wrong terms like $2y^4$ and $-6y^3$. PT-MoE maintains information completeness throughout the solution process and avoids hallucinations, ultimately producing the correct polynomial expression. Notably, PT-MoE achieves this with a more concise solution process, demonstrating efficient problem-solving steps while maintaining accuracy. 

\subsection{Ablation Studies}

To comprehensively evaluate the design choices in PT-MoE, we conduct ablation studies on \textbf{five influencing variables}: soft prompt length, trainable parameters, number of experts, routing mechanisms, and model size. For each variable, we keep other variables fixed at their default values (soft prompt length=40, trainable parameters$\approx$80K, number of experts=2, probationary-selective routing, 1B base model) while varying the target component to identify its influence on model performance.

\begin{figure*}[t!]
\centering
\begin{subfigure}[b]{0.225\textwidth}
    \centering
    \includegraphics[width=\textwidth]{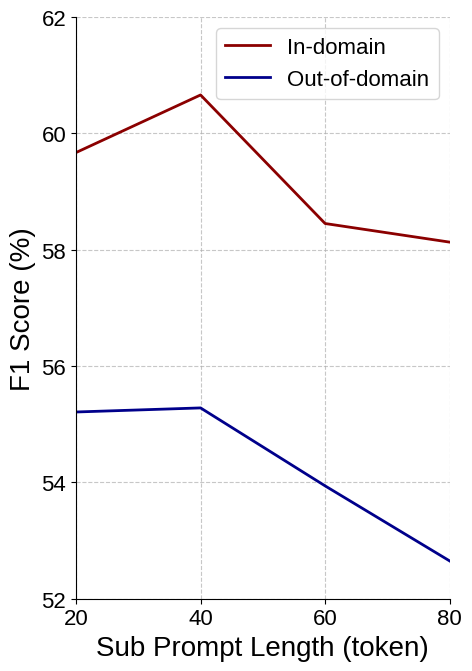}
    \label{as-1}
\end{subfigure}
\begin{subfigure}[b]{0.225\textwidth}
    \centering
    \includegraphics[width=\textwidth]{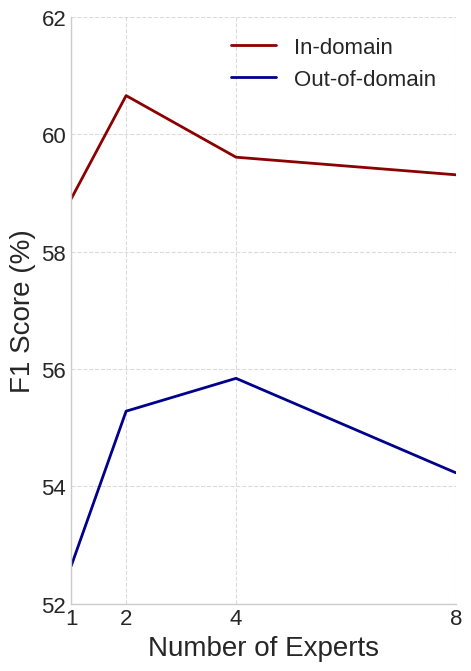}
    \label{as-2}
\end{subfigure}
\begin{subfigure}[b]{0.225\textwidth}
    \centering
    \includegraphics[width=\textwidth]{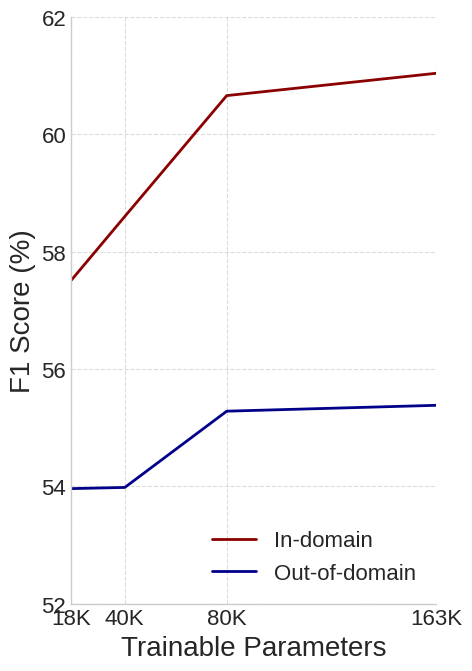}
    \label{as-3}
\end{subfigure}
\begin{subfigure}[b]{0.225\textwidth}
    \centering
    \includegraphics[width=\textwidth]{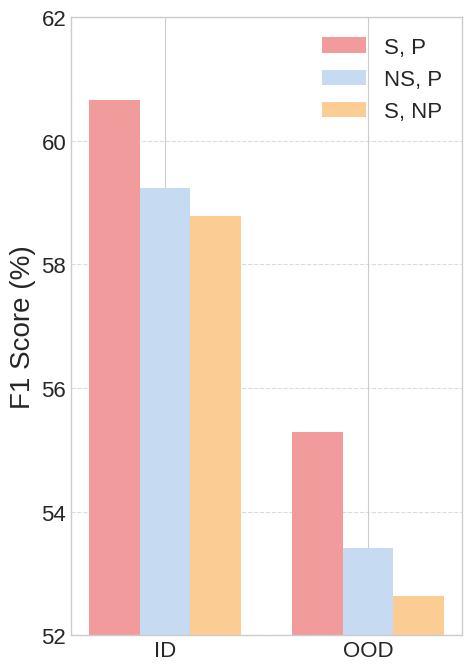}
    \label{as-4}
\end{subfigure}
\\
\vspace{-10pt}
\caption{\textbf{Ablation studies} on key components of PT-MoE, showing the influence of (Left) prompt length, (Center left) number of experts, (Center right) trainable parameters, and (Right) routing mechanisms ((N)S: (Non-)Selective, (N)P: (Non-)Probationary) on in-domain (ID) and out-of-domain (OOD) performance.}
\label{as}
\end{figure*}

\textbf{Soft prompt length.} We evaluate prompt lengths ranging from 20 to 80 tokens (Figure \ref{as} Left). Three consistent observations appear: (1) In-domain performance exceeds out-of-domain across all lengths, maintaining a 5-6\% F1 score margin; (2) Both domains achieve optimal performance at 40 tokens, with peak F1 scores of 60.66\% and 55.28\% respectively; and (3) Performance in both domains follows a similar trend, improving up to 40 tokens then decreasing. These findings indicate that the optimal prompt length is domain-agnostic, though the absolute performance levels remain domain-dependent.

\textbf{Number of experts.} We investigate the influence of expert count by varying it from 1 to 8 (Figure \ref{as} Center left). There are three key points: (1) Single-expert configuration yields the poorest performance (58.90\% and 52.64\% F1 for in-domain and out-of-domain), demonstrating the necessity of MoE; (2) Performance exhibits an initial increase followed by decrease, with in-domain peaking at N=2 (60.66\% F1) and out-of-domain at N=4 (55.84\% F1), suggesting different optimal routing abilities for each domain; (3) In-domain tasks consistently outperform out-of-domain scenarios by a 4-6\% F1 margin across all expert counts. These observations demonstrate that the optimal number of experts varies by domain type and highlight the importance of balancing expert focus with routing difficulty.

\textbf{Trainable parameters.} We vary the parameter count from 18K to 163K to analyze its influence on model performance (Figure \ref{as} Center Right). Three key observations appear: (1) Performance consistently improves with increasing parameters, from 57.51\% to 61.04\% F1 for in-domain and 53.96\% to 55.38\% F1 for out-of-domain tasks, and notably maintains stability without decrease even at higher parameter counts, differing from conventional prompt tuning methods; (2) While both in-domain and out-of-domain tasks show increasing trend, they exhibit different parameter dependence behaviours, in-domain tasks demonstrate rapid improvement before 80K parameters, while out-of-domain tasks show accelerated growth in the 40K-80K range; (3) In-domain performance maintains a consistent advantage over out-of-domain tasks across all parameter ranges, with F1 scores differing by approximately 4-6\%. These findings show that PT-MoE effectively leverages additional parameters to achieve continuous performance gains.

\textbf{Routing mechanisms.} We examine two key routing design choices (Figure \ref{as} Right): selective routing, which uses only the highest-weighted expert versus non-selective routing that utilizes all experts with their respective weights, and probationary routing, which multiplies the output by the router's selection probability versus non-probationary routing that uses original outputs. Our results show four key findings: (1) The combination of selective and probationary routing (S, P) consistently outperforms other configurations (NS, P and S, NP) across both in-domain (60.66\% vs 59.24\% and 58.78\% F1) and out-of-domain tasks (55.28\% vs 53.41\% and 52.64\% F1), suggesting the complementary benefits of focused expert utilization and confidence-based output; (2) Probationary routing demonstrates superior performance over its non-probationary counterpart, indicating the value of incorporating router confidence in the final output; (3) Under probationary conditions, selective routing achieves 1.42\% higher F1 score while reducing utilized parameters compared to non-selective routing, highlighting the effectiveness and efficiency of domain-specific knowledge; (4) All routing configurations maintain higher performance on in-domain tasks compared to out-of-domain scenarios, though the relative performance rankings remain consistent across domains. These findings collectively demonstrate that the selective probationary routing mechanism achieves an optimal balance between model performance and computational efficiency.

\begin{table}[t!]
\centering
\small
\begin{tabular}{c@{\hspace{1cm}}c@{\hspace{1cm}}c@{\hspace{1cm}}c}
\hline
 & \textbf{PT} & \textbf{SMoP} & \textbf{PT-MoE} \\
\hline
GSM8K & 56.70 & \textbf{61.78} & 59.74 \\
SVAMP & 69.36 & \textbf{74.69} & 72.81 \\
ASDIV & 76.41 & 80.06 & \textbf{81.39} \\
MAWPS & 70.70 & 70.70 & \textbf{78.02} \\
MP500 & 59.00 & 60.80 & \textbf{63.60} \\
\hline
Average & 66.43 & 69.61 & \textbf{71.11} \\
\hline
\end{tabular}
\caption{Performance comparison (accuracy \%) of standard and MoE-based prompt tuning methods on mathematical problem solving tasks using a \textbf{3B} base model.}
\label{math-3b}
\end{table}

\textbf{Model size.} We conduct additional studies using a 3B version of the base model, comparing PT-MoE with PT and the MoE-integrated method, SMoP (Table \ref{math-3b}). Three key findings are found: (1) PT-MoE maintains its advantage at larger sizes, achieving the highest average accuracy of 71.11\%, outperforming standard PT (66.43\%) and SMoP (69.61\%). (2) SMoP shows size-dependent behaviour: while underperforming PT on the 1B model (56.77\% vs 56.25\%), it outperforms PT on the 3B model (69.61\% vs 66.43\%). (3) PT-MoE demonstrates robust performance by outperforming the baselines on three out of five mathematical datasets. These findings collectively validate the size-independence and stability of PT-MoE across different model sizes.

\subsection{Efficiency Analysis}

\begin{figure}[t!]
    \centering
    \includegraphics[width=0.45\textwidth]{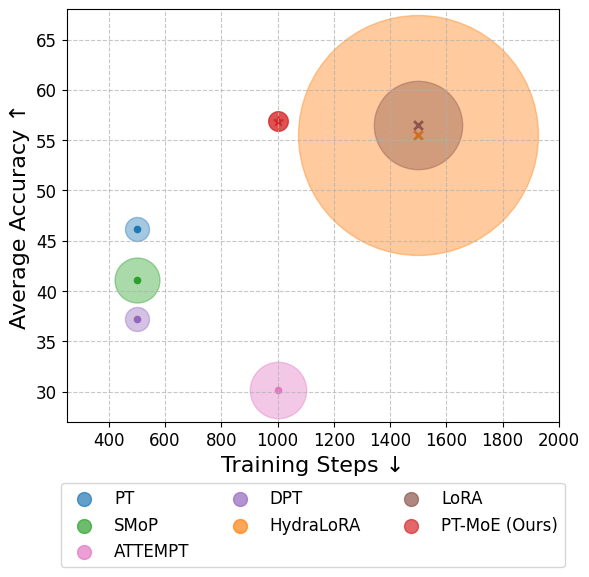}
    \vspace{-5pt}
    \caption{Parameter and training \textbf{efficiency comparison} across different methods. The x-axis shows training steps for the highest performance after training parameter search, while the y-axis shows the average accuracy on math datasets. Circle sizes indicate the number of trainable parameters, with larger circles indicating more parameters.}
    \label{efficiency}
\end{figure}

Results in Figure \ref{efficiency} demonstrate PT-MoE's efficiency across both computational and parametric aspects. PT-MoE achieves the highest performance with only moderate training steps and minimal parameters (80k). In contrast, LoRA and HydraLoRA require more parameters and training steps to achieve comparable performance. Other prompt tuning methods such as PT, SMoP, and DPT converge fast but achieve lower performance, suggesting a potential trade-off between training efficiency and model effectiveness. These results validate that PT-MoE balances the computational cost, parameter efficiency, and model performance.

\section{Conclusions}
\label{Conclusions}

This work introduces PT-MoE, a novel parameter-efficient framework that integrates matrix decomposition with MoE routing for prompt tuning. Our results across 17 datasets demonstrate that PT-MoE achieves state-of-the-art performance while maintaining parameter efficiency, outperforming existing methods in both QA and mathematical tasks. Through ablation studies, we identify optimal configurations for prompt length, expert count, and routing mechanisms, providing insights for future parameter-efficient tuning approaches. 

Future directions include exploring hierarchical routing mechanisms to better deal with diverse task distributions, and extending PT-MoE to continual learning scenarios for efficient adaptation and knowledge transfer across tasks.

\section*{Limitations}
\label{Limitations}
While PT-MoE shows promising results, there are some key points that need to be noted. Like other fine-tuning methods, people should be careful of training data licensing and usage rights. Furthermore, while our results demonstrate strong performance across benchmark tasks, developing comprehensive insights for diverse applications would benefit from broader community contributions and open-source collaboration.

\section*{Ethics Statement}
No ethical approval was required for this study.

\section*{Availability Statement}
The codes and models related to this paper are uploaded to the open-source community at https://github.com/ZongqianLi/PT-MoE.

\bibliography{custom}

\appendix

\section{Appendix}
\label{sec:appendix}

\subsection{Implementation Details}

We provide implementation details, including training hyperparameters in Table \ref{training-parameters}, inference parameters in Table \ref{inference-parameters}, and method-specific configurations in Tables \ref{method-parameters} to facilitate reproducibility. The models were finetuned on one node with four A100 80G.

\begin{table}[th!]
    \centering
    \small
    \begin{tabular}{ccc}
        \hline
        {} & \textbf{QA} & \textbf{Math} \\
        \hline
        max\_new\_tokens & 100 & 768 \\
        num\_beams & 1 & 1 \\
        do\_sample & False & False \\
        temperature & 1.0 & 1.0 \\
        top\_p & 1.0 & 1.0 \\
        pad\_token\_id & pad\_token\_id & pad\_token\_id \\
        eos\_token\_id & eos\_token\_id & eos\_token\_id \\
        early\_stopping & True & True \\
        length\_penalty & 1.0 & 1.0 \\
        \hline
    \end{tabular}
    \caption{\textbf{Inference parameters} for QA and mathematical tasks.}
    \label{inference-parameters}
\end{table}

\begin{table*}[th!]
    \centering
    \small
    \begin{tabular}{ccc}
        \hline
        {} & \textbf{QA} & \textbf{Math} \\
        \hline
        Train steps & \{500, 1000, 1500\} for PT-based methods & \{500, 1000, 1500\} \\
          & \{200, 600, 1000\} for LoRA-based methods & \\
        Optimizer & AdamW & AdamW \\
        Max length & 512 & 768 \\
        warmup\_steps & 500 & 500 \\
        learning\_rate & 2e-5 & 2e-5 \\
        per\_device\_train\_batch\_size & 32 & 16 \\
        lr\_scheduler\_type & constant\_with\_warmup & constant\_with\_warmup \\
        gradient\_accumulation\_steps & 2 & 2 \\
        \hline
    \end{tabular}
    \caption{\textbf{Training hyperparameters} for QA and mathematical tasks. \{\} means parameter search.}
    \label{training-parameters}
\end{table*}

\begin{table*}[th!]
    \centering
    \small
    \begin{tabular}{p{0.1\textwidth}p{0.7\textwidth}}
        \hline
        \textbf{Method} & \textbf{Details} \\
        \hline
        LoRA & r=1; lora\_alpha=16; target\_modules=["q\_proj", "v\_proj"]; lora\_dropout=0; bias="none"; task\_type=TaskType.CAUSAL\_LM \\
        HydraLoRA & r=1; alpha=16; target\_modules=["q\_proj", "v\_proj"]; dropout=0.0; num\_b\_matrices=2; Router: nn.Sequential(nn.Linear(input\_dim, num\_b\_matrices)); Initialization: A: nn.init.kaiming\_uniform\_(, a=math.sqrt(5)), B: nn.init.zeros\_() \\
        PT & \multicolumn{1}{p{0.7\textwidth}}{Soft prompt length: 40; Initialization: Specific words} \\
        DPT & \multicolumn{1}{p{0.7\textwidth}}{Soft prompt length: 40; low\_rank\_dim = 39; Initialization: Specific words; Decomposition method: SVD} \\
        SMoP & \multicolumn{1}{p{0.7\textwidth}}{Total soft prompt length: 40; Number of experts: 2; Initialization: Specific words; Noise: *(1+torch.randn\_like()*0.01)} \\
        ATTEMPT & \multicolumn{1}{p{0.7\textwidth}}{Total soft prompt length: 40; Number of experts: 2; Encoder: nn.Linear(embedding\_dim, projection\_dim=1), nn.Linear(projection\_dim=1, embedding\_dim), nn.LayerNorm(embedding\_dim); Initialization: Specific words}  \\
        PT-MoE & Soft prompt length: 40; Number of expert: 2; Rank: 36; Router: nn.Linear(embedding\_dim, num\_prompts); Noise: *(1+torch.randn\_like()*0.01) \\
        \hline
    \end{tabular}
    \caption{\textbf{Method configurations} for various PEFT methods.}
    \label{method-parameters}
\end{table*}

Prompt for MRQA:

\begin{quote}
\begin{scriptsize}
\begin{lstlisting}[breaklines=true, basicstyle=\ttfamily, numbers=none, breakindent=0pt, xleftmargin=0pt, ]
<|start_header_id|>user<|end_header_id|>\n\nExtract the exact text span from the given context that directly answers the question, without modifying or combining multiple parts of the text.\n\nContext: {}\n\nQuestion: {}<|eot_id|><|start_header_id|>assistant<|end_header_id|>\n\nAnswer:
\end{lstlisting}
\end{scriptsize}
\end{quote}

Prompt for Math datasets:

\begin{quote}
\begin{scriptsize}
\begin{lstlisting}[breaklines=true, basicstyle=\ttfamily, numbers=none, breakindent=0pt, xleftmargin=0pt, ]
<|start_header_id|>user<|end_header_id|>\n\nSolve the question and your response should end with \"The answer is: [answer]\".\n\nQuestion: {}<|eot_id|><|start_header_id|>assistant<|end_header_id|>\n\nAnswer: 
\end{lstlisting}
\end{scriptsize}
\end{quote}

Texts used to initialize soft prompt for finetuning on MRQA:

\begin{quote}
\begin{scriptsize}
\begin{lstlisting}[breaklines=true, basicstyle=\ttfamily, numbers=none, breakindent=0pt, xleftmargin=0pt, ]
(
    "Read the following context and answer the question. "
    "Extract the answer from the context. "
    "The answer is a span of the context."
    "Answer the question directly."
    "Use the original words in the context."
    "Do not introduce any words not present in the context."
)
\end{lstlisting}
\end{scriptsize}
\end{quote}

Texts used to initialize soft prompt for finetuning on Math datasets:

\begin{quote}
\begin{scriptsize}
\begin{lstlisting}[breaklines=true, basicstyle=\ttfamily, numbers=none, breakindent=0pt, xleftmargin=0pt, ]
(
    "Read the question carefully and make sure you understand it before beginning. "
    "Pay close attention to the details and requirements of the question. "
    "Answer the question, ensuring your response is relevant to what is asked. "
    "Ensure your answer is both accurate and correct."
)
\end{lstlisting}
\end{scriptsize}
\end{quote}

\subsection{Environment}

\begin{verbatim}
    python==3.11.5
    torch==2.3.1+cu118
    transformers==4.46.0
    datasets==2.18.0
    huggingface_hub==0.24.2
    deepspeed==0.15.3
    wandb==0.14.2
    numpy==1.23.5
    tqdm==4.66.4
\end{verbatim}

\end{document}